\documentclass{article}

\usepackage{microtype}
\usepackage{graphicx}
\usepackage{subcaption}
\usepackage{booktabs}
\usepackage{hyperref}
\usepackage{needspace}

\usepackage[preprint]{icml2026}

\usepackage{amsmath}
\usepackage{amssymb}
\usepackage{mathtools}
\usepackage{amsthm}

\usepackage[capitalize,noabbrev]{cleveref}

\theoremstyle{plain}

\theoremstyle{definition}

\theoremstyle{remark}

\icmltitlerunning{Randomized Masked Finetuning for PII Mitigation in LLMs}

\begin{document}

\twocolumn[
  \icmltitle{Randomized Masked Finetuning: An Efficient Way to Mitigate\\
    Memorization of PIIs in LLMs}

  \icmlsetsymbol{equal}{*}

  \begin{icmlauthorlist}
    \icmlauthor{Kunj Joshi}{institution1}
    \icmlauthor{David A. Smith}{institution1}
  \end{icmlauthorlist}

  \icmlaffiliation{institution1}{Khoury College of Computer Sciences, Northeastern University, Boston, MA, USA}

  \icmlcorrespondingauthor{Kunj Joshi}{joshi.kun@northeastern.edu}

  \icmlkeywords{Machine Learning, Privacy, Large Language Models, Memorization, PII Protection}

  \vskip 0.3in
]

\printAffiliationsAndNotice{}

\begin{abstract}
 Language models are known to memorize training data, which they can later regurgitate when prompted in specific ways. Memorization may result in severe privacy concerns, especially when training datasets contain sensitive data like personally identifiable information (PII). We introduce Randomized Masked Fine-Tuning (RMFT), a novel privacy-preserving fine-tuning technique that reduces PII memorization while minimizing performance impact. Using the Enron Email Dataset, we demonstrate that RMFT achieves an 80.81\% reduction in Total Extraction Rate and 80.17\% reduction in Seen Extraction Rate compared to baseline fine-tuning, outperforming deduplication methods while maintaining only a 5.73\% increase in perplexity. We present MaxTER, a Pareto-optimal evaluation framework for assessing privacy-utility tradeoffs, and show the performance of RMFT vs Deduplication by Area Under The Response Curve (AURC) metric.
\end{abstract}

\section{Introduction}

Language models are known to memorize training data, which they can later regurgitate when prompted in specific ways \citep{carlini2021extractingtrainingdatalarge}. Memorization may result in severe privacy concerns, especially when training datasets contain sensitive data like personally identifiable information (PII) \citep{yang2024memorization}. Prior studies show that memorization in language models is largely driven by data duplication \citep{lee2022deduplicating, pmlr-v162-kandpal22a}—token sequences that appear multiple times in the dataset tend to be memorized more often than sequences appearing once or infrequently \citep{kandpal-etal-2024-user, 10.1162/TACL.a.49}. Therefore, we move forward with the assumption that a \emph{higher number of duplicates for a specific PII increases the risk of memorization.}

Keeping this in mind, we devised a novel fine-tuning technique that reduces the number of times a particular PII appears in the dataset while keeping the original PII intact for future downstream tasks. This technique is called \emph{Randomized Masked Fine-Tuning} (RMFT). In our study, we focus on email addresses as the PII of interest. RMFT identifies all occurrences of each email in the dataset and preserves only the first occurrence in its original form. All subsequent duplicate occurrences are masked by generating structurally similar email addresses using randomly selected components—firstnames, lastnames, and domains—that closely mimic the original email's format. Using this method, we reduce individual email occurrences from hundreds or thousands down to a single appearance.

We fine-tune the model with this modified dataset and evaluate memorization using an extractable memorization setup, where we prompt the model with the same set of randomly sampled prompts used for the baseline fine-tuned model to ensure uniformity across experiments. Our results demonstrate that RMFT is highly effective in reducing memorization: we achieve an 80.81\% reduction in Total Extraction Rate (TER) and an 80.17\% reduction in Seen Extraction Rate (SER) compared to baseline fine-tuning with no dataset modifications. This substantial privacy improvement comes with minimal impact on model performance—RMFT incurs only a 5.73\% increase in perplexity as measured on the Enron Testing Split.Validation on GPT-Neo-1.3B demonstrates similar results (80.42\% TER reduction, 7.08\% perplexity increase), confirming 
generalizability across architectures (Appendix~\ref{app:gpt_neo}).

To benchmark RMFT's effectiveness, we adapt the deduplication methodology from \citet{lee2022deduplicating} to our PII-specific context by removing datapoints containing duplicate email addresses rather than duplicate text sequences. We then conduct a comprehensive evaluation comparing RMFT and deduplication across three dimensions: email extraction rates, model performance, and training efficiency. Our results demonstrate that RMFT achieves better privacy-utility tradeoff as compared to deduplication. However, deduplication is computationally more efficient, requiring less preprocessing time than RMFT's masking procedure. 

\needspace{6\baselineskip}

\subsection{Contributions}

This study introduces:
\begin{itemize}
\item \textbf{Randomized Masked Fine-tuning (RMFT)}: A novel, privacy-preserving fine-tuning technique designed to minimize performance impact while significantly reducing PII memorization.
\item \textbf{Novel Privacy-Utility Evaluation Framework}: We introduce \textbf{MaxTER}, a Pareto-optimal analysis framework that jointly characterizes the privacy-utility tradeoff in memorization mitigation techniques. Adapting extraction methodologies from \citet{carlini2021extractingtrainingdatalarge} and \citet{nasr2023scalableextractiontrainingdata}, we define Total Extraction Rate (TER) and Seen Extraction Rate (SER) as privacy metrics (Section~4.2.1), and Mean Delta Perplexity (MDP) as our utility metric (Section~4.2.2). For a given perplexity threshold $\tau$, MaxTER identifies the checkpoint that achieves maximum TER reduction, enabling practitioners to select optimal privacy-utility operating points. We quantify overall framework efficiency using Area Under the Response Curve (AURC) to systematically compare RMFT and deduplication across privacy-utility tradeoff spaces.

\end{itemize}

\section{Related Work}

\subsection{PII Memorization and Leakage}
Large language models have been shown to memorize and leak personally identifiable information (PII) from training data, with successful extraction demonstrated on publicly available models like GPT-2 \citep{carlini2021extractingtrainingdatalarge, nasr2023scalableextractiontrainingdata, huang-etal-2022-large}. Beyond individual PII extraction, research has characterized how memorization manifests throughout training, identifying patterns such as assisted, retained, forgotten, and immediate memorization \citep{borkar2025privacyrippleeffectsadding}. Studies have also examined how models reproduce token sequences, phrases, and entire passages verbatim \citep{hartmann2023sokmemorizationgeneralpurposelarge, wang2025generalizationvsmemorizationtracing}.

\subsection{Mitigation Techniques}
Current approaches to reducing memorization include unlearning—removing specific data post-training \citep{chen2023unlearnwantforgetefficient, si2023knowledgeunlearningllmstasks}—and deduplication, which removes repeated sequences before training \citep{kandpal-etal-2024-user, lee2022deduplicating}. However, unlearning can be inefficient and may not fully eliminate memorization \citep{to2025harrypotterhereprobing}. Alternative techniques include differential privacy for membership inference protection \citep{pmlr-v162-kandpal22a}, model architecture modifications \citep{sakarvadia2025mitigatingmemorizationlanguagemodels}, and activation steering on privacy-sensitive neurons \citep{suri2025mitigatingmemorizationllmsusing}.

\subsection{Deduplication as a Privacy-Preserving Technique}

Among mitigation techniques, deduplication has emerged as particularly effective and practical for reducing memorization. \citet{lee2022deduplicating} demonstrated that exact deduplication—removing training sequences appearing verbatim multiple times—substantially reduces memorization rates while improving model performance on downstream tasks. Extensions beyond exact matching, such as semantic deduplication \citep{abbas2023semdedupdataefficientlearningwebscale}, demonstrate that removing semantically similar sequences can further reduce memorization while preserving dataset diversity. \citet{tirumala2022memorizationoverfittinganalyzingtraining} confirmed that deduplication consistently reduces memorization throughout training dynamics without sacrificing model capability, while \citet{pmlr-v162-kandpal22a} established the underlying causal mechanism: token sequences appearing more frequently in training data exhibit proportionally higher extraction rates. These findings collectively establish deduplication as a principled and effective approach to mitigating memorization.

However, standard deduplication operates at the document or sequence level—typically removing passages of 50+ tokens that appear multiple times—and does not specifically target structured PII such as email addresses, which may be embedded within otherwise unique contexts. This gap motivates our work: we adapt deduplication to PII-aware scenarios by removing datapoints containing duplicate email addresses, and we introduce RMFT as an alternative that reduces duplication while preserving dataset size and linguistic diversity. Given deduplication's established effectiveness and its conceptual similarity to our approach, we use the adapted version as explained in \ref{sec:deduplication} as our primary baseline for comparison.

\subsection{Privacy-Utility Tradeoffs}
Balancing privacy preservation with model utility remains a central challenge in LLM deployment. Recent work has evaluated various privacy-preserving techniques \citep{10.1145/3696410.3714531, das2025revisitingprivacyutilityefficiency}, with promising results showing that anonymization can reduce privacy leakage by 97-99\% while incurring only 10\% performance degradation \citep{pasch-cha-2025-balancing}. However, existing studies typically evaluate privacy and utility in isolation. Our work addresses this gap by proposing a unified framework that jointly characterizes privacy-utility tradeoffs for PII mitigation techniques.

\section{Dataset Used}

We utilize a raw version of the \textbf{Enron Email Dataset}~\cite{shetty2004enron} to construct our training, validation, and testing splits. The dataset is divided into three partitions following a 100:10:1 ratio---the training set contains 10,000 datapoints, the validation set 1,000 datapoints, and the testing set 100 datapoints, all randomly sampled. Each datapoint corresponds to an individual email sample containing one or more personally identifiable email addresses.

To evaluate the two key aspects of our approach---\textbf{memorization} and \textbf{performance}---the experiment is structured into two evaluation phases:

\begin{itemize}
\item \textbf{Memorization Evaluation via Generation:} We employ a set of 16,000 English prompts sampled from Common Crawl (CC-MAIN-2024-10, February 2024 snapshot). Prompts were extracted from the WARC file using a fixed random seed to ensure reproducibility and represent realistic, open-ended user queries rather than explicitly targeted extraction prompts. For generation, we use the following configuration: prompts are truncated to 50 tokens as input, greedy decoding, and a maximum generation length of 256 tokens (including the prompt). This setup ensures deterministic, reproducible extraction experiments across all checkpoints and training techniques. Email addresses are extracted from generated text using regex pattern matching, with matches validated against the original Enron dataset. Example prompts are provided in Appendix~\ref{app:prompts}.

\item \textbf{Perplexity Evaluation:} To assess language modeling performance, we use the testing subset of the Enron dataset. This provides an accurate measure of how fine-tuned models generalize to unseen yet domain-consistent data.
\end{itemize}

For randomized masking, we construct a small, curated list of commonly used domain names and top-level domains (TLDs), publicly available on our GitHub repository (Anonymous, 2026). Each list contains approximately 15 examples, used to generate realistic but anonymized email variants during masking.

\subsection{Exploratory Data Analysis}

Before applying any dataset modifications, we conducted an exploratory data analysis (EDA) on the Enron Email dataset to better understand the statistical distribution of personally identifiable email addresses. Specifically, we aimed to quantify how often certain PIIs recur across the dataset, since our hypothesis---consistent with \citet{kandpal-etal-2024-user}---is that frequent token repetition correlates with higher memorization likelihood in LLMs.

\begin{figure}[t]
    \centering
    \includegraphics[width=0.7\columnwidth]{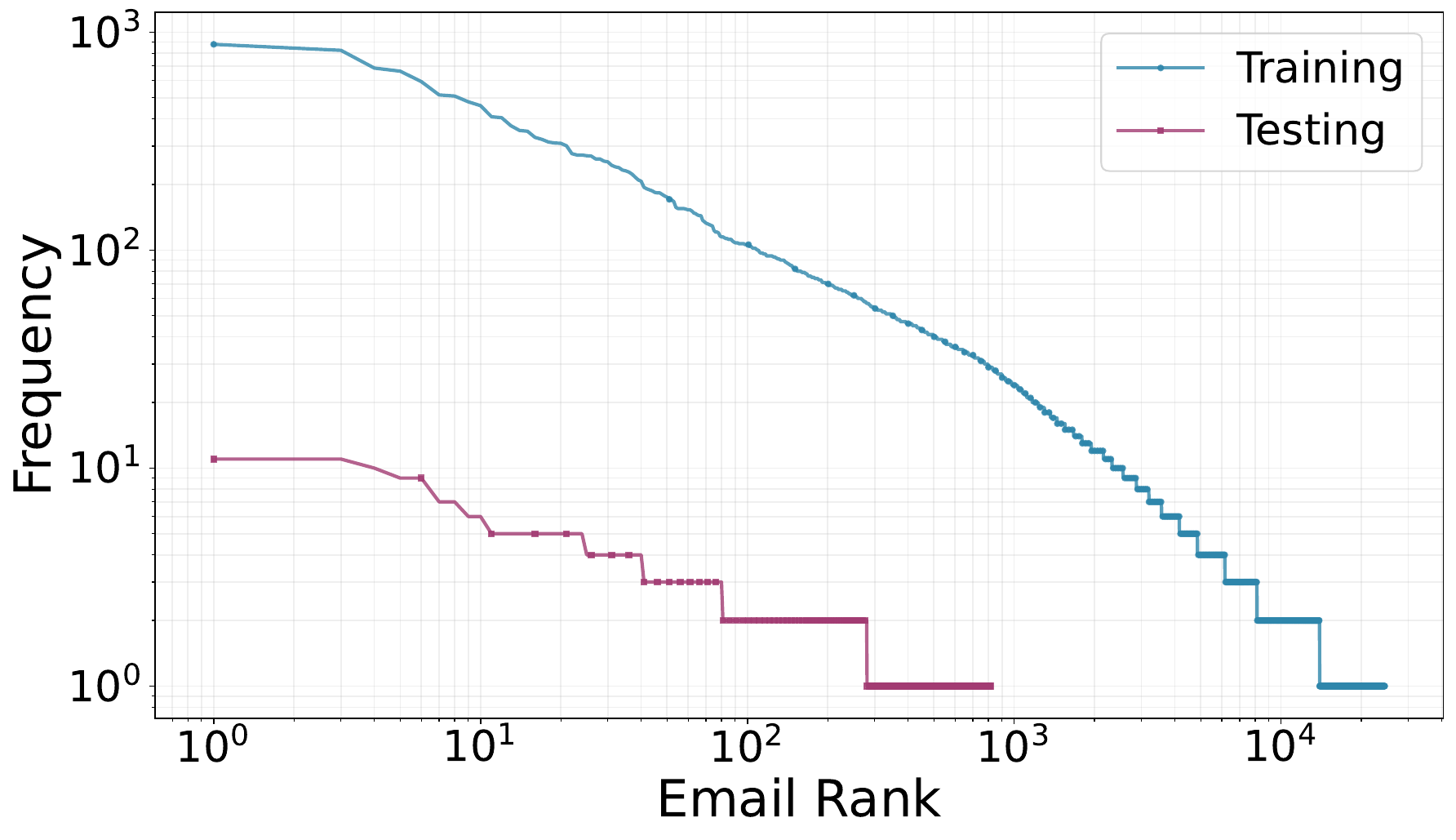}
    \caption{Email address frequency distributions on a log-log scale. Both training and testing sets exhibit power-law (heavy-tailed) distributions, with a small number of emails appearing hundreds of times while most appear once or twice. The most frequent email in the training set appears 879 times (rank 1), while the most frequent in the testing set appears 11 times. This severe duplication in training data motivates RMFT's approach to reduce memorization risk by masking duplicate occurrences.}
    \label{fig:loglog_plot}
\end{figure}

\Cref{fig:loglog_plot} shows the frequency distribution on a log-log scale, revealing power-law (heavy-tailed) behavior in both training and testing sets. In the training set, the most frequent email appears 879 times among 10,000 datapoints, with frequencies declining smoothly across 24,480 unique email addresses. Such high-frequency occurrences are particularly problematic for privacy, as repeated exposure during fine-tuning can lead models to memorize entire token sequences corresponding to these email addresses. 

This severe duplication directly motivated our two dataset-level interventions: (1) Randomized Masked Fine-Tuning (RMFT), which reduces recurrence through randomized replacements while preserving structure; and (2) Deduplication, which explicitly removes repeated headers to prevent repeated exposure to the same email tokens.

\section{Methodology}

We perform experiments to test the effectiveness of randomized masking in preventing PII memorization. As we have taken our inspiration from Kandpal results on N-Gram deduplication, we also compare our results with an adapted version of deduplication for PIIs. \Cref{fig:methodology} depicts a summarized version of methodologies used in randomized masked fine-tuning and deduplication.


Our experiment compares randomized masked fine-tuning with deduplication. We have created a suite of evaluations that considers both memorization and performance of trained models. We use metrics defined in \cref{sec:metrics} to compare memorization and performance of the fine-tuning techniques.

Over both phases we are fine-tuning our model for 3 epochs, checkpointing the LLM at every 10\% training steps, utilizing a fixed learning rate of 0.00002 over all three training methods, as in \citet{borkar2025privacyrippleeffectsadding}. We also save the data seen until every checkpoint and compare checkpoint-based metrics for the different fine-tuning techniques. For generation experiments, greedy decoding with no sampling is used, setting the temperature to 0, top\_p to 1.0, top\_k to 0, and deterministic beam search switched off. The average TER, average SER, and average mean delta perplexity are calculated as the average respective values over the 30 checkpoints.

\subsection{Deduplication vs RMFT}

For the experiment comparing deduplication and RMFT, we evaluate on three criteria: memorization, model performance, and training performance. For training performance, time complexity is considered.

\subsubsection{Randomized Masking}

As defined in the introduction, RMFT masks multiple recurring PIIs with a randomized, structured PII that closely mimics the original PII. Using this methodology, we reduce the occurrences of certain PIIs from thousands down to a single occurrence in the dataset.

This is done using an indexed table of email addresses and their occurrences over the datapoints, their starting index and ending index. Each of these occurrences is sorted based on a combination of datapoint index and starting index. The first occurrence is left untouched, but the rest of the occurrences are masked with a randomized PII. For this experiment we utilize email addresses as the PII. Our randomized creation of emails is as follows:

\begin{enumerate}
\item We create a set of first names, last names, domain names, and TLDs as they appear in the set of originally available emails.
\item For each email to be masked, an anchor is decided randomly. The anchor is a part of the email address that remains unchanged.
\item We select the remaining two parts of the email address randomly from the collected sets of parts in Step 1.
\item The email address is created with the selected parts and anchor. The newly generated email replaces the original email at the exact position where it occurred.
\end{enumerate}

For example, if the original email is \texttt{kay.mann@enron.com}, we might randomly select the last name as the anchor and replace the remaining components with sampled parts from the collected sets. This could yield \texttt{bob.mann@yahoo.org}, preserving structural consistency (\texttt{FIRST.LAST@DOMAIN.TLD}) while preventing exact memorization.

\subsubsection{Deduplication Training}
\label{sec:deduplication}
We adapt deduplication for PII-aware context, closely following the structure defined by \citet{lee2022deduplicating}. They define deduplication as dropping datapoints that contain token sequences of 400 tokens or more that are repeated multiple times. We adapt it to PII-aware context, where we drop datapoints that have a repeating email address.

Instead of discarding entire datapoints, we remove only the headers containing repeated email addresses, retaining the main email text. This preserves natural language context while still eliminating recurring PIIs from the structural header format. Such selective deduplication provides a more balanced tradeoff between privacy and linguistic richness.

\subsection{Metrics Used}
\label{sec:metrics}

We adapt extraction rate methodologies from \citet{carlini2021extractingtrainingdatalarge} and \citet{nasr2023scalableextractiontrainingdata} to the PII context, defining two complementary metrics that measure memorization at different granularities: Total Extraction Rate (TER) for global PII extraction across the entire dataset, and Seen Extraction Rate (SER) for checkpoint-level PII extraction relative to training exposure. \citet{borkar2025privacyrippleeffectsadding} similarly measure email address extraction rates over training sequences, which informed our metric design. For performance assessment, we utilize Mean Delta Perplexity (MDP) to compare training techniques.

\subsubsection{Memorization Metrics}

Total Extraction Rate is defined as the number of email addresses leaked at a checkpoint divided by the total number of email addresses available in the original set. Leaked email addresses are extracted from generated text using regex pattern matching for the format \texttt{[first\_name].[last\_name]@[domain].[tld]}, with matches validated against the original email set.
TER is calculated as:
\begin{equation}
TER_{ckpt_i} = \left( \frac{|LeakSet_{ckpt_i}|}{|OG\text{-}EMAIL\text{-}SET|} \right) \times 100
\end{equation}

Seen Extraction Rate is defined as the number of email addresses leaked at a checkpoint divided by the total number of emails seen until that checkpoint. To calculate the total number of emails seen until checkpoint $i$, we take the union of emails seen at checkpoint $i$ with the set of emails seen before checkpoint $i$.

Seen Extraction Rate is calculated as:
\begin{equation}
SER_{ckpt_i} = \left( \frac{|LeakSet_{ckpt_i}|}{|SeenSet_{ckpt_i} \cup SeenSet_{ckpt_{i-1}}|} \right) \times 100
\end{equation}

Both TER and SER play an important role in determining memorization within the checkpoints of a given training technique. Higher SER and TER values mean a higher rate of emails are being memorized. Lower levels of TER and SER are preferred for better privacy.

\subsubsection{Performance Metrics}

We utilize perplexity on a fixed set of prompts for all checkpoints of the three training techniques to calculate the performance of each. We use the WikiText2 benchmark testing set, which contains around 2,500 prompts.

As the variance in calculated perplexities can be high, we utilize Mean Delta Perplexity (MDP), which measures the increase in perplexity compared to baseline.

Mean Delta Perplexity is calculated as:
\begin{equation}
\Delta PP_{ckpt_k} = \sum_{i=1}^{N} \left( (PP_{ckpt_k}^{TM})_{p_i} - (PP_{ckpt_k}^{BL})_{p_i} \right) \times \frac{1}{N}
\end{equation}

This calculates the average difference in perplexity per prompt for a given checkpoint in two different training methods. For this experiment, we associate a higher MDP with worse performance, considering baseline perplexity as our standard threshold.

\section{Results and Observations}

In this section, we summarize the key quantitative outcomes of our experiments across three fine-tuning techniques: baseline fine-tuning, randomized masked fine-tuning (RMFT), and deduplication training. Metrics are averaged over 30 checkpoints unless otherwise stated. \Cref{tab:summary} summarizes the memorization and perplexity observations.

\begin{table}[t]
\caption{Summary of observations over three different training techniques. Each value represents an average over all checkpoints. Extraction rates are percentages. The best performing technique in each column is bolded.}
\label{tab:summary}
\centering
\small
\begin{tabular}{@{}lccc@{}}
\toprule
Training & PPL & TER & SER \\
Technique & ($\downarrow$) & ($\downarrow$) & ($\downarrow$) \\
\midrule
Baseline & \textbf{6.432} & 0.308 & 0.342 \\
RMFT & 6.798 & \textbf{0.049} & \textbf{0.059} \\
Deduplication & 7.963 & 0.084 & 0.102 \\
\bottomrule
\end{tabular}
\end{table}

\subsection{Extraction Rates}

Both RMFT and deduplication significantly reduce memorization relative to baseline fine-tuning. RMFT achieves an 80.81\% reduction in Total Extraction Rate (TER) and an 80.17\% reduction in Seen Extraction Rate (SER) compared to the baseline, outperforming deduplication's 69.38\% TER and 68.84\% SER reductions. This demonstrates RMFT's superior ability to mitigate PII memorization while maintaining dataset utility.

At the final checkpoint, RMFT achieves an 87.62\% reduction in TER and 86.0\% in SER, compared to deduplication's 74.22\% and 72.9\%, respectively. These results suggest that RMFT continues to improve privacy robustness over training progression. \Cref{fig:extraction_rates} shows the Total Extraction Rates and Seen Extraction Rates for all three training techniques on a fixed web crawl dataset of 16,000 prompts.

\begin{figure}[t]
\centering
\begin{subfigure}{0.48\textwidth}
    \centering
    \includegraphics[width=\textwidth]{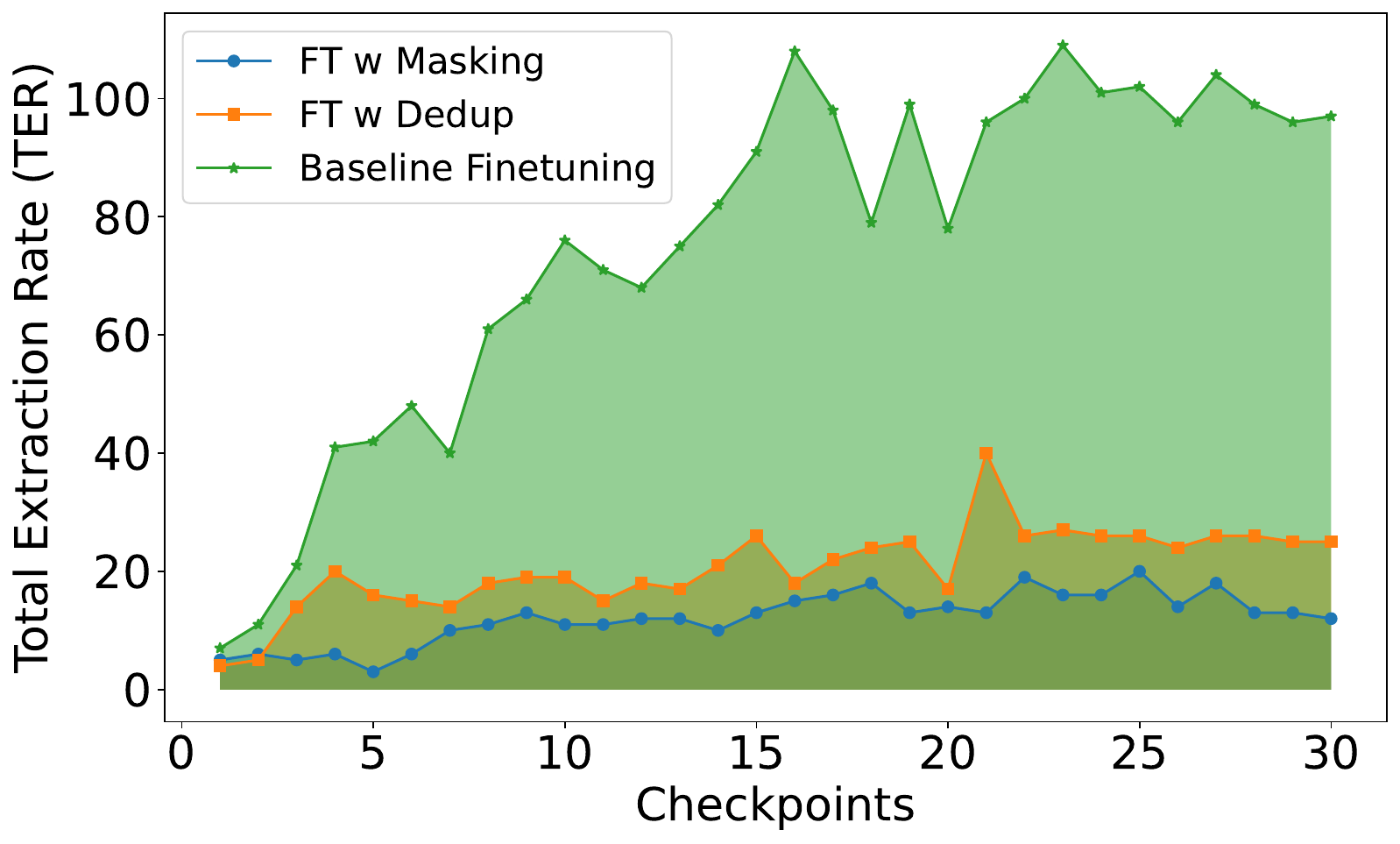}
    \caption{Total Extraction Rate (TER)}
    \label{fig:ter}
\end{subfigure}
\hfill
\begin{subfigure}{0.48\textwidth}
    \centering
    \includegraphics[width=\textwidth]{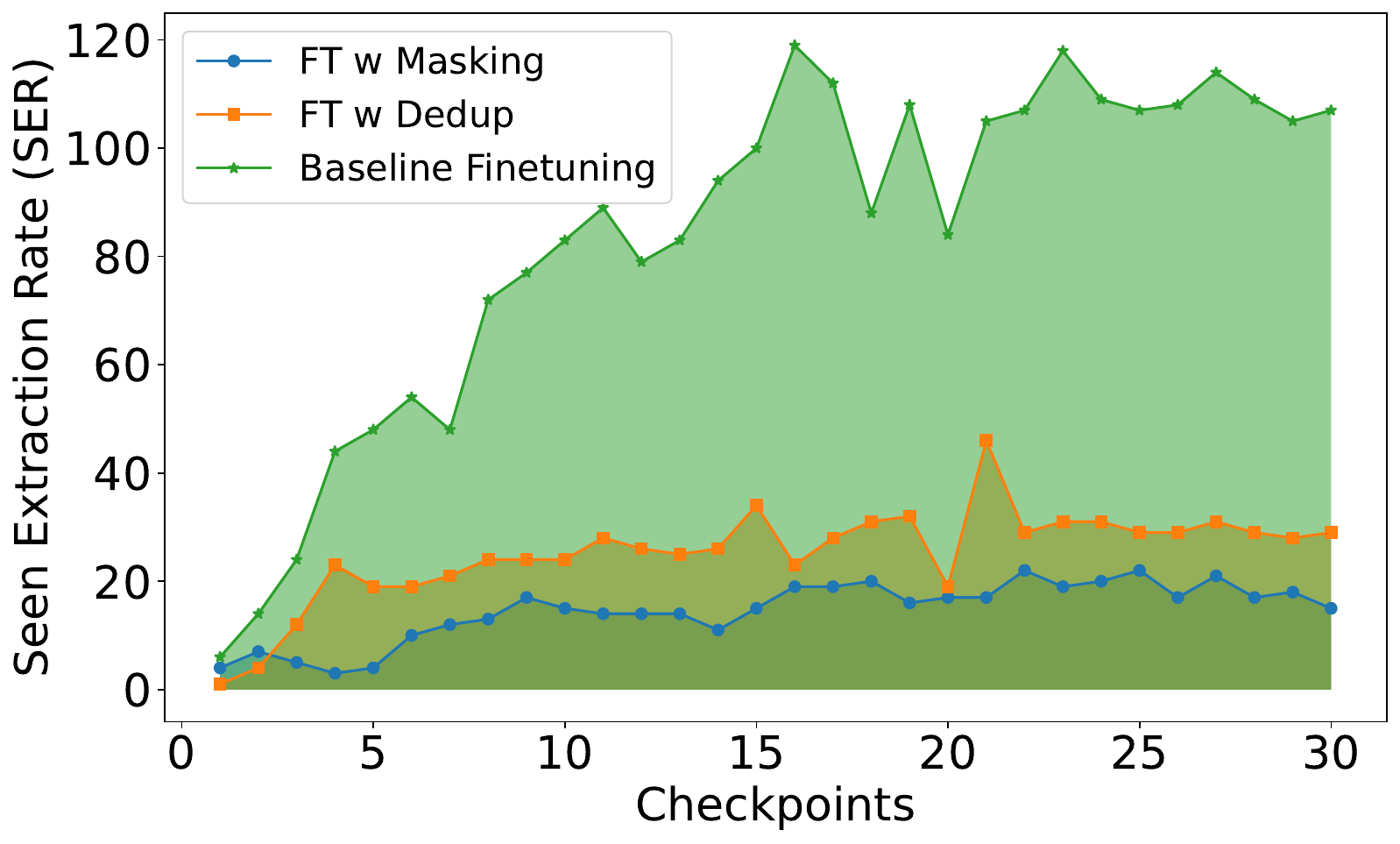}
    \caption{Seen Extraction Rate (SER)}
    \label{fig:ser}
\end{subfigure}
\caption{Total Extraction Rate (TER) and Seen Extraction Rate (SER) across training checkpoints for all three fine-tuning techniques. Lower values indicate fewer PIIs memorized by the model.}
\label{fig:extraction_rates}
\end{figure}

\subsection{Mean Delta Perplexity}

\begin{figure}[t]
\centering
\begin{subfigure}{0.48\textwidth}
    \centering
    \includegraphics[width=\textwidth]{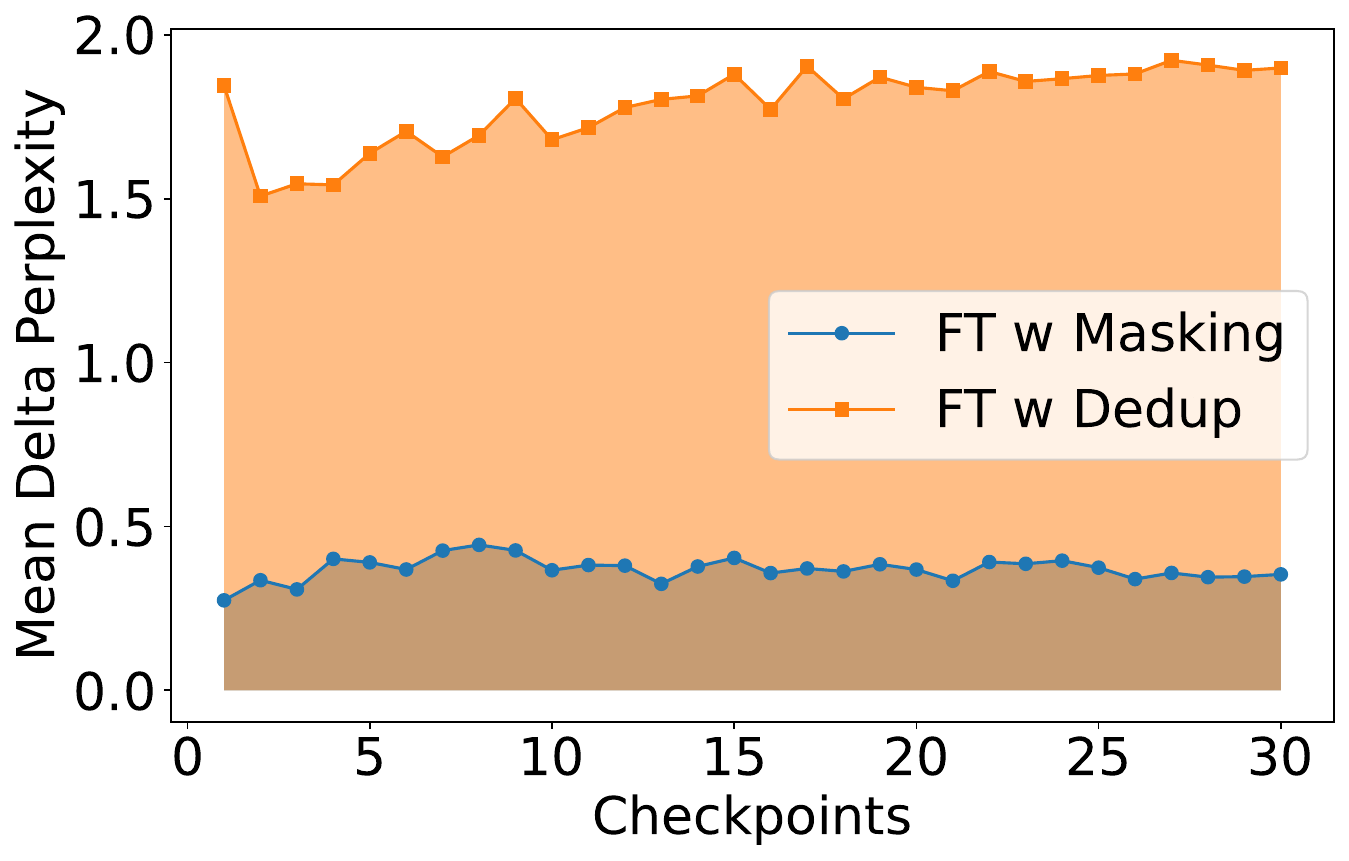}
    \caption{Mean Delta Perplexity}
    \label{fig:mdp}
\end{subfigure}
\hfill
\begin{subfigure}{0.48\textwidth}
    \centering
    \includegraphics[width=\textwidth]{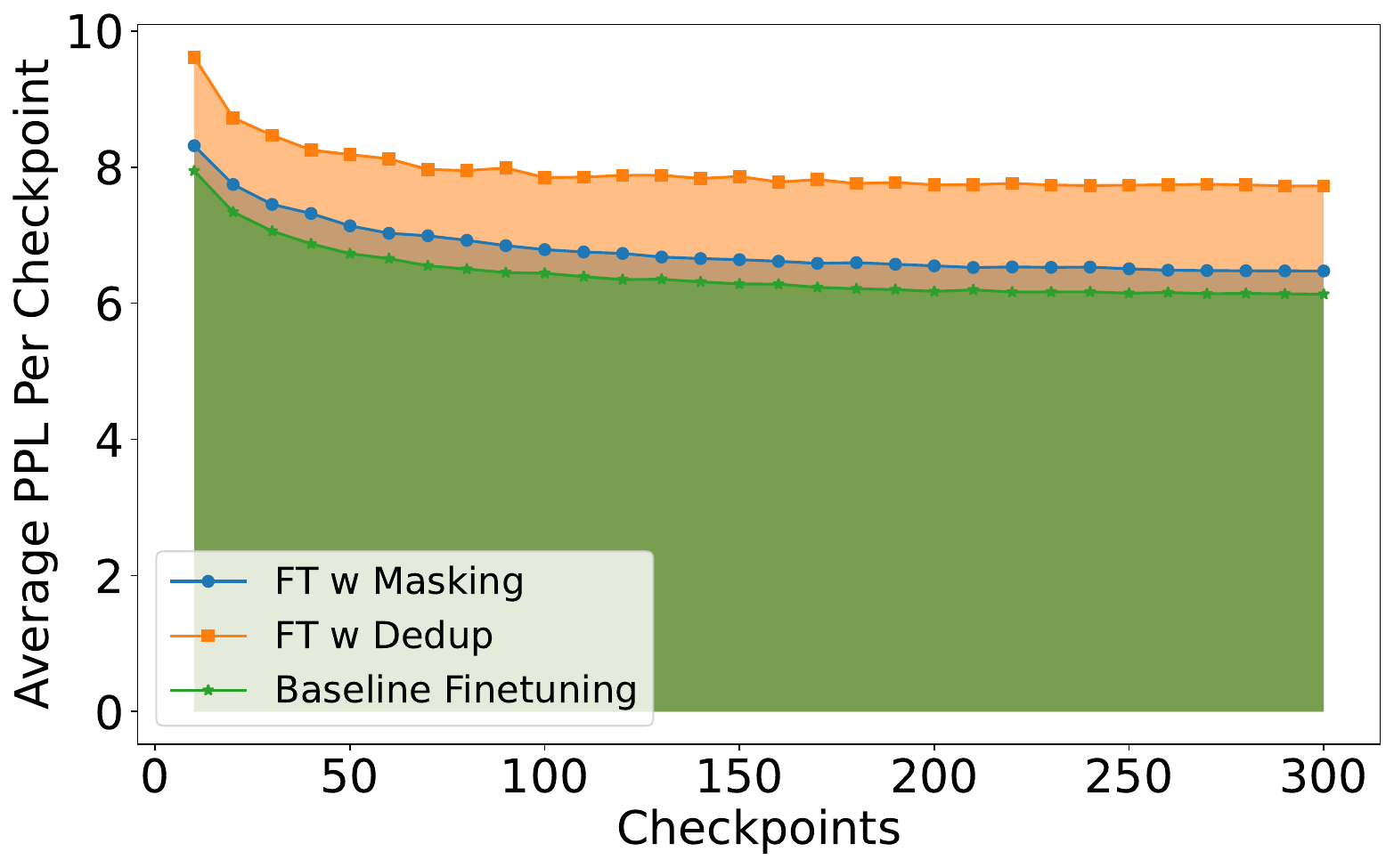}
    \caption{Average Perplexity per Checkpoint}
    \label{fig:avgppl}
\end{subfigure}
\caption{(a) Mean Delta Perplexity of Deduplication vs Randomized Masked Finetuning (RMFT). Lower MDP indicates less deviation from baseline performance. (b) Average Perplexity per Checkpoint for all training techniques. This curve conforms the Mean Delta Perplexity of RMFT being lower than Deduplication}
\label{fig:extraction_rates}
\end{figure}

The Mean Delta Perplexity (MDP) curves exhibit a consistent pattern: RMFT has consistently less degradation in performance compared to deduplication. RMFT achieves a significantly lower average perplexity increase (+5.73\%) relative to the baseline, compared to deduplication's +23.85\%. RMFT maintains 14.5\% lower average perplexity on all checkpoints than deduplication, confirming reduced performance degradation. The final training checkpoint for RMFT has a 5.56\% increase in perplexity compared to 26\% increase in deduplication. \Cref{fig:mdp} shows the Mean Delta Perplexity plot for both RMFT and deduplication at each checkpoint.

\subsection{Time Complexity Analysis}

We calculate the time complexity of training and compare the two techniques---deduplication and RMFT. We only consider the time complexity of preprocessing, as both training techniques follow the same training steps once the data is preprocessed.

We assume $N$ datapoints for training, each datapoint has on average $k$ PIIs appearing, the length of each datapoint is $l$, and the number of unique emails is $u$. The total number of PIIs occurring in the whole dataset would be $N \cdot k$. For our experiments, $N = 10{,}000$, $k \approx 6$, and $u = 24{,}480$.

\subsubsection{Randomized Masked Finetuning}

The time complexity for randomized masked fine-tuning is $O(N \cdot k \cdot |\log u|)$. The $|\log u|$ term arises because we sort the PIIs based on their appearance. This is important as we want to keep the first appearance untouched but need to mask the rest.

\subsubsection{Deduplication}

Our modified deduplication approach is divided into two phases: deduplication of headers and recombination of headers. The deduplication of headers takes $O(Nk)$ time. The recombination process takes $O(N \log N)$ time, and hence we report the dominating term as the final time complexity: $O(N \log N)$.

For our study, we took 10,000 training datapoints, each having 6 emails on average and 24,480 unique email addresses. RMFT takes 15 minutes and 25 seconds of preprocessing on average, whereas our modified deduplication takes 10 minutes and 12 seconds of preprocessing on an NVIDIA H200 GPU. This clearly indicates that deduplication is more efficient than randomized masked fine-tuning for preprocessing time.

\section{Analysis}

We perform a Pareto frontier analysis on our experimental results to identify the most optimal training checkpoint for a given percentage threshold increase in Mean Delta Perplexity (MDP) that yields the maximum percentage reduction in Total Extraction Rate (TER). We adopt TER percentage drop rather than SER percentage drop for this analysis, since the Pareto frontier captures a global optimal tradeoff between privacy and performance. TER reflects a global measure of overall memorization across the dataset, whereas SER reflects localized memorization behavior limited to the model's seen subset.

We define a MaxTER function that returns the maximum TER percentage drop for a given threshold of MDP:

\begin{equation}
MaxTER_t(\tau) = \max_{\{i \mid MDP_{i,t} \leq \tau\}} \Delta TER_{i,t}
\end{equation}

where $\tau$ is the MDP threshold acceptable, $i$ is the checkpoint number, and $t$ is the training technique.

Percentage MDP is calculated as:
\begin{equation}
MDP_{i,t}\% = \left( \frac{PPL_{i,t} - PPL_{i,b}}{PPL_{i,b}} \right) \times 100
\end{equation}

Percentage TER Drop is calculated as:
\begin{equation}
\Delta TER_{i,t}\% = \frac{TER_{i,t} - TER_{i,b}}{TER_{i,b}} \times 100
\end{equation}

\begin{figure}[t]
\centering
\includegraphics[width=\columnwidth, height=2.5in, keepaspectratio]{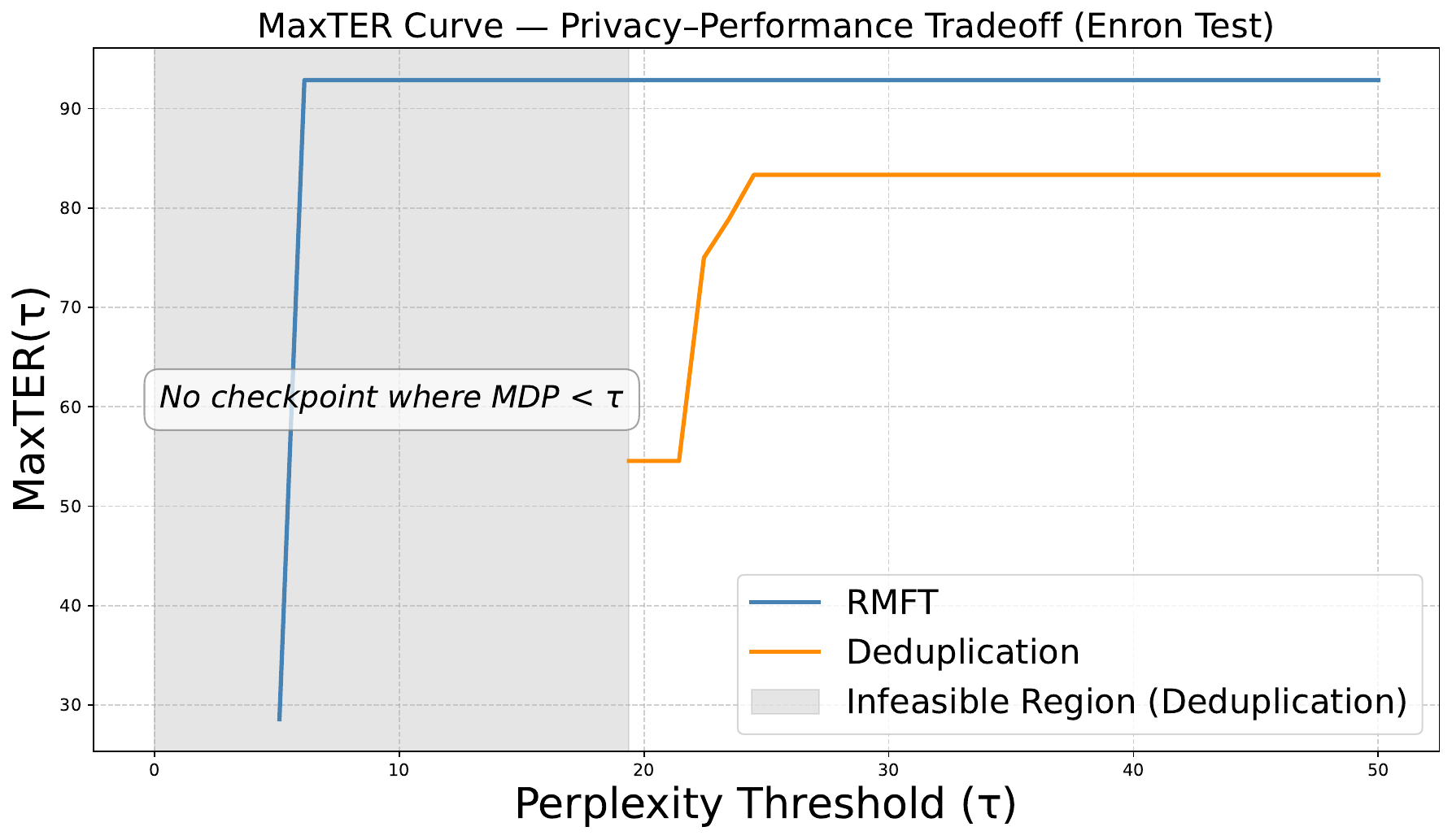}
\caption{MaxTER analysis identifying checkpoints that optimize the tradeoff between TER reduction and perplexity increase. RMFT outperforms deduplication by achieving greater TER reductions while incurring lower perplexity penalties. No deduplication checkpoint satisfies the constraint of perplexity increase below 20\% (threshold $\tau$), representing the grey infeasible region}
\label{fig:pareto}
\end{figure}

\Cref{fig:pareto} presents the MaxTER analysis comparing deduplication and RMFT. Notably, deduplication exhibits no feasible checkpoints below a 20\% increase in MDP, marking the region prior to this threshold as infeasible. In contrast, RMFT demonstrates consistent performance across all MDP thresholds, achieving higher reductions in TER for equivalent or lower increases in perplexity. We compute the Area Under the Response Curve (AURC) to quantify overall privacy-performance efficiency. RMFT achieves an AURC of approximately 92.52 units², substantially surpassing deduplication's 17.09 units².

This analysis addresses how to evaluate the tradeoff between reduced memorization and potential degradation in model performance. The proposed MaxTER function provides an effective framework for identifying the optimal checkpoint for each fine-tuning technique by quantifying the balance between reduced memorization (percentage drop in TER) and increased model perplexity (percentage rise in MDP). RMFT consistently outperforms deduplication in memorization reduction, achieving lower TER (0.049 vs 0.084) and SER (0.059 vs 0.102) across all evaluated conditions. On in-distribution evaluation using the Enron held-out test set, RMFT achieves substantially superior AURC (92.52 vs 17.09 units²), demonstrating its effectiveness when deployment conditions match training distributions---the primary use case for privacy-preserving fine-tuning.

\section{Conclusion}

In conclusion, this study demonstrates that Randomized Masked Fine-tuning (RMFT) is an effective and computationally efficient approach to mitigating the memorization of personally identifiable information (PIIs) in large language models. Compared to traditional deduplication, RMFT achieves substantially greater reductions in Total and Seen Extraction Rates while maintaining lower degradation in model performance as measured by Mean Delta Perplexity. Through MaxTER analysis, we show that RMFT consistently provides a more favorable tradeoff between privacy preservation and performance retention. Our results establish RMFT as a promising, scalable fine-tuning strategy for enhancing privacy robustness in LLMs when deployment distributions match training conditions without compromising their generative quality---ideal for privacy-preserving training.

\section{Limitations and Future Work}

While RMFT demonstrates strong potential as a privacy-preserving fine-tuning strategy, several limitations remain. First, our experiments are restricted to the fine-tuning stage of model training, whereas the RMFT framework can, in principle, be applied to pretraining as well. The effect of randomized masking during large-scale pretraining---where token frequency and contextual learning dynamics differ significantly---remains unexplored and may yield different outcomes in terms of both privacy and performance.

We note that on out-of-distribution evaluation (WikiText2, Web Crawl), 
deduplication achieves higher AURC than RMFT (Appendix~\ref{app:ood_eval}). This 
occurs because RMFT is optimized for in-distribution privacy-preserving 
fine-tuning scenarios where deployment conditions match training 
distributions.  Deduplication's broader text-level approach may generalize better to 
unrelated domains, while RMFT's PII-specific design excels at protecting 
sensitive information in the target domain.

Finally, RMFT's current success is closely tied to the structured nature of PIIs such as email addresses, which can be effectively masked with realistic, randomized replacements. Extending this method to mitigate memorization of arbitrary token sequences---where semantic and syntactic coherence must be preserved---poses a greater challenge. This limitation also motivated our adaptation of deduplication into a PII-aware variant, as standard deduplication better addresses generic token-level repetition.

Our main evaluation focuses on GPT2-XL, with additional evaluation on GPT-Neo 1.3B (Appendix ~\ref{app:gpt_neo}) demonstrate consistent and comparable performance. While the methodology is architecture-agnostic, validation across diverse model families (e.g., Llama, Qwen, Gemma, OLMo) would strengthen generalizability and is the focus of ongoing work.

\section*{Acknowledgements}

We thank the reviewers for their valuable feedback.

\section*{Impact Statement}

This paper presents work whose goal is to advance the field of machine learning with respect to privacy protection in large language models. There are many potential societal consequences of our work, particularly in enabling safer deployment of LLMs in privacy-sensitive domains such as healthcare, finance, and legal services. By reducing PII memorization, our approach can help mitigate privacy risks while maintaining model utility, contributing to more responsible AI development practices.

\bibliography{references}
\bibliographystyle{icml2026}

\newpage
\appendix
\onecolumn

\section{Example of Randomized PII Creation}

\begin{figure}[h]
\centering
\includegraphics[width=0.9\textwidth, height=4in, keepaspectratio]{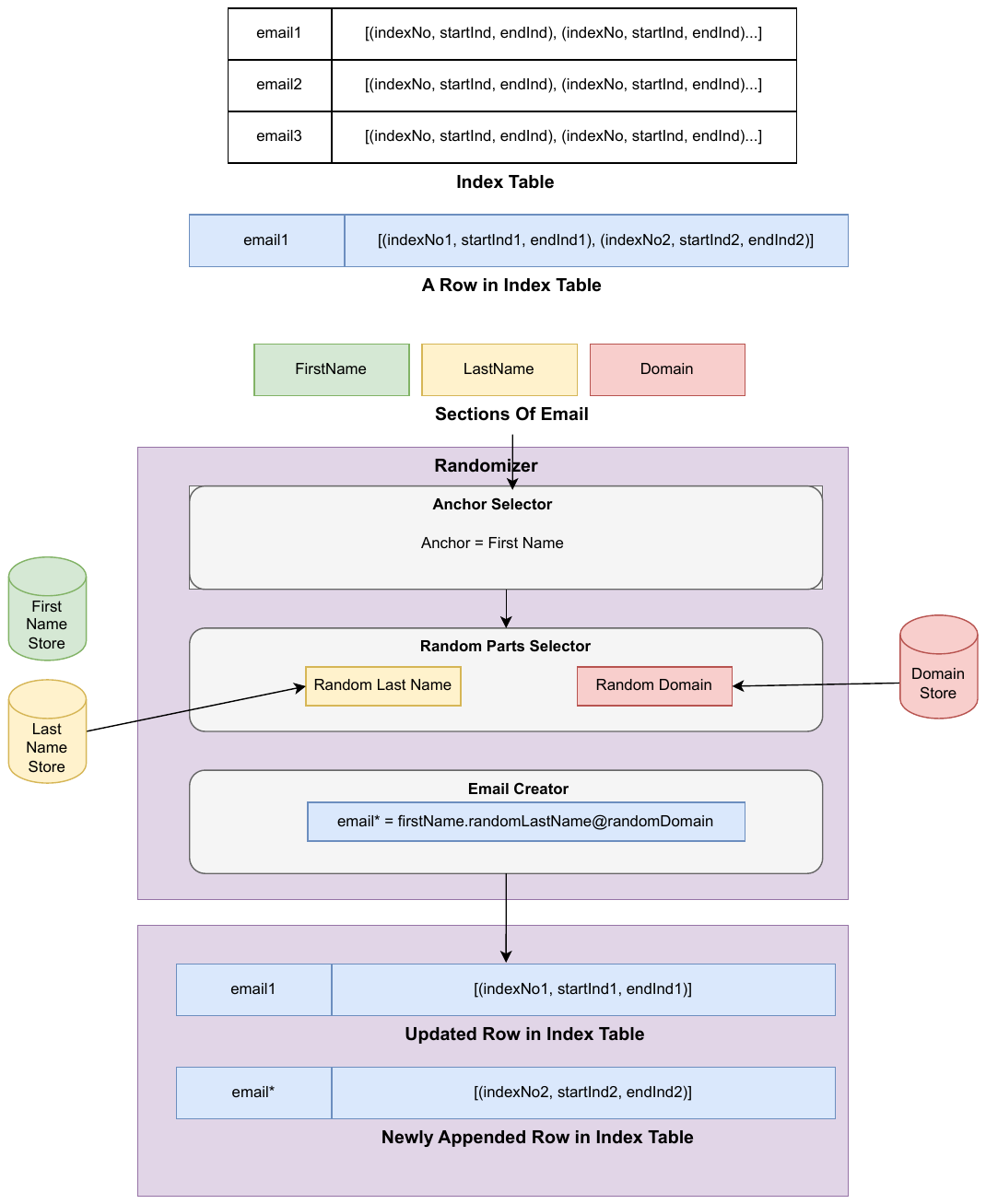}
\caption{This diagram shows an example of how RMFT replaces an occurrence of an email with a randomized email. The randomizer selects an anchor (first name, last name, or domain) and keeps it fixed while selecting the remaining parts randomly from the store.}
\label{fig:appendix_randomization}
\end{figure}

\Cref{fig:appendix_randomization} shows how the randomizer in RMFT creates new email addresses. The index table at the top is created using the original dataset. The first name store and last name store are populated using the already existing emails in the dataset. The domain store contains popularly used domains.

\subsection{Example Dataset}

Consider the following email:

\begin{verbatim}
Message-ID: <33099599.1075845686859.JavaMail.evans@thyme>
Date: Wed, 4 Apr 2001 02:32:00 -0700 (PDT)
From: kay.mann@enron.com
To: suzanne.adams@att.net
Subject: Re: Wednesday

How's everything coming up? Did Warren help you last night?

Kay
\end{verbatim}

Let's take the example of \texttt{suzanne.adams@att.net}:

\begin{enumerate}
\item The three facets of the email are: suzanne (first name), adams (last name), and att.net (domain).
\item The randomizer selects an anchor randomly, say first name.
\item The randomizer selects random parts for last name and domain, say watterberg and google.com.
\item The new email would be \texttt{suzanne.watterberg@google.com}.
\end{enumerate}

\section{MaxTER Evaluation on Different Datasets}
\label{app:ood_eval}
We present evaluations of RMFT versus deduplication via the MaxTER function on three datasets: Enron held-out test dataset, WikiText2 test dataset ~\cite{radford2019language}, and randomly collected web crawl prompt dataset. \Cref{tab:datasets} summarizes all results.
\begin{figure}[h]
\centering
\begin{subfigure}{0.32\textwidth}
    \centering
    \includegraphics[width=\textwidth]{pareto_enron_test}
    \caption{Enron Held-Out Dataset}
    \label{fig:maxter_enron}
\end{subfigure}
\hfill
\begin{subfigure}{0.32\textwidth}
    \centering
    \includegraphics[width=\textwidth]{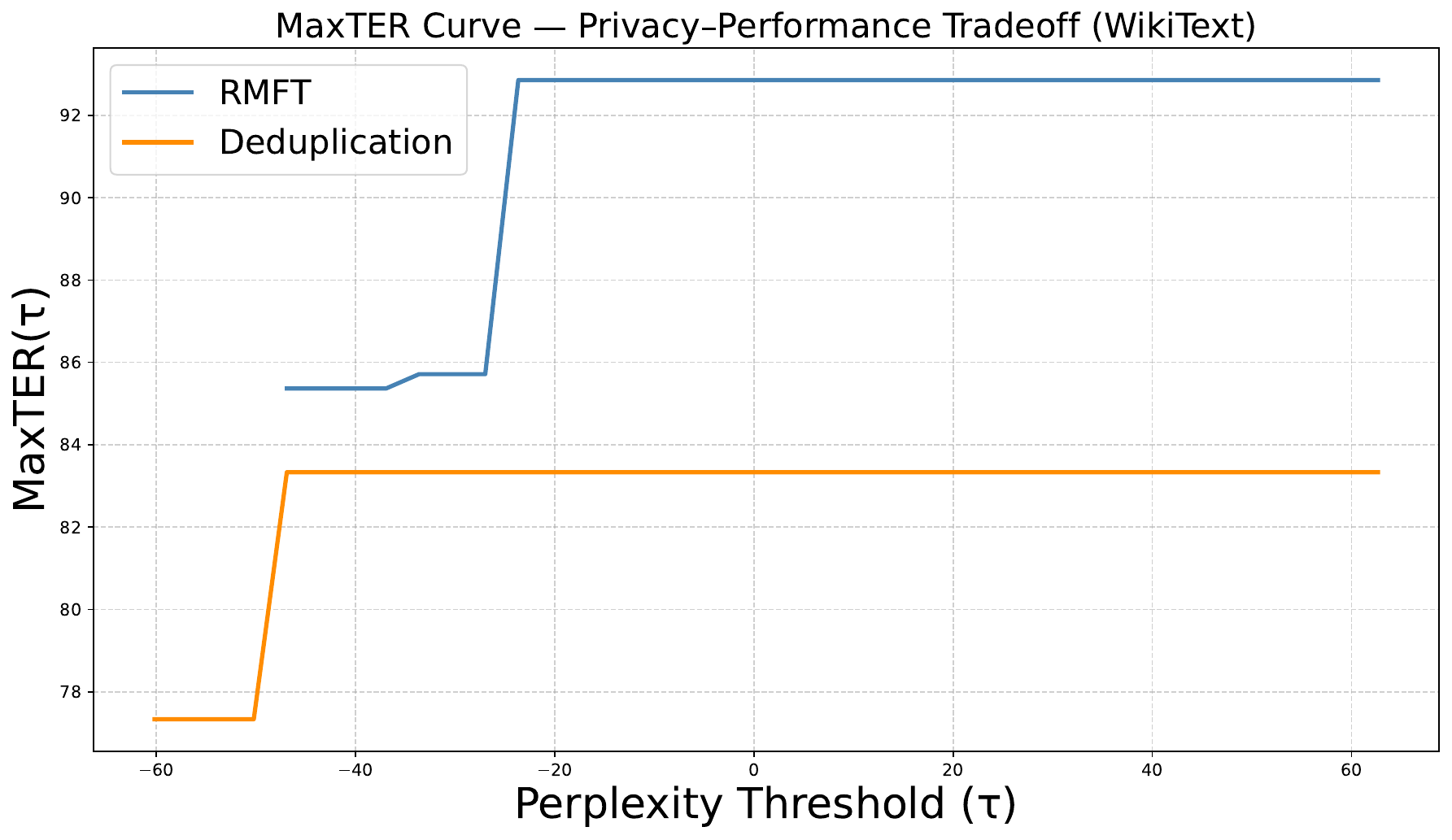}
    \caption{WikiText2 Test Dataset}
    \label{fig:maxter_wiki}
\end{subfigure}
\hfill
\begin{subfigure}{0.32\textwidth}
    \centering
    \includegraphics[width=\textwidth]{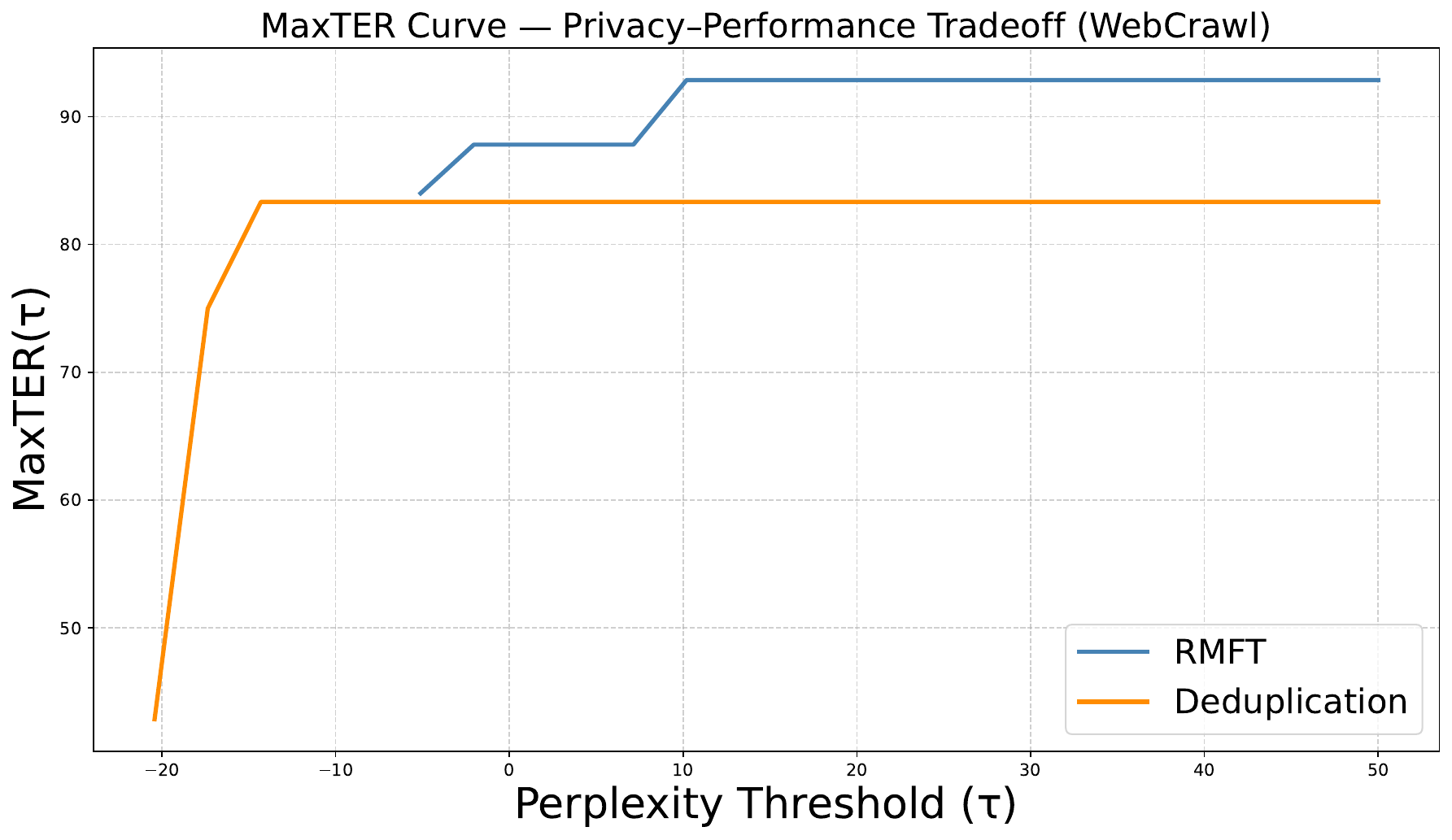}
    \caption{Web Crawl Prompts}
    \label{fig:maxter_web}
\end{subfigure}
\caption{MaxTER curves for RMFT vs Deduplication across three different datasets. The curves show the privacy-performance tradeoff measured by Area Under the Response Curve (AURC).}
\label{fig:maxter_comparison}
\end{figure}

\begin{table}[h]
\caption{Comparison of training techniques across three different datasets. MaxTER AURC represents the direct relation between memorization reduction and performance.}
\label{tab:datasets}
\centering
\small
\begin{tabular}{@{}lcccccc@{}}
\toprule
& \multicolumn{2}{c}{Enron Held Out} & \multicolumn{2}{c}{WikiText2} & \multicolumn{2}{c}{Web Crawl} \\
\cmidrule(lr){2-3} \cmidrule(lr){4-5} \cmidrule(lr){6-7}
Technique & Avg PPL & AURC & Avg PPL & AURC & Avg PPL & AURC \\
\midrule
Baseline & 6.432 & - & 605.23 & - & 2168.39 & - \\
RMFT & 6.798 & 92.52 & 585.43 & 75.548 & 2214.70 & 76.84 \\
Dedup & 7.963 & 17.09 & 531.97 & 82.081 & 1847.56 & 82.09 \\
\bottomrule
\end{tabular}
\end{table}

From the above table we can see how deduplication has a higher MaxTER AURC for different datasets such as WikiText2 test and web crawl prompts, whereas RMFT maintains a higher MaxTER curve for Enron held-out dataset. RMFT heavily outperforms deduplication in the Enron held-out dataset, whereas the performance on WikiText2 is comparable between both techniques. Deduplication outperforms RMFT on randomly sampled prompts from web crawl.

\section{Generalization to GPT-Neo-1.3B}
\label{app:gpt_neo}

To validate RMFT's generalizability across model architectures, we repeated our experiments on GPT-Neo-1.3B, which differs from GPT-2-XL in both size (1.3B vs 1.5B parameters) and architecture details. We use identical experimental setup, training hyperparameters, and evaluation methodology as described in Section 4.

\subsection{Results and Observations}

Table~\ref{tab:gptneo-comparison} compares RMFT performance across GPT-2-XL and GPT-Neo-1.3B. Metrics are averaged over 30 checkpoints unless otherwise stated. Extraction rates are percentages.

\begin{table}[h]
\centering
\caption{Comparison of RMFT performance on GPT-2-XL vs GPT-Neo-1.3B. Both models show consistent memorization reduction with minimal performance impact.}
\label{tab:gptneo-comparison}
\begin{tabular}{@{}lcccc@{}}
\toprule
Model & Avg TER & Avg SER & Avg PPL & PPL \\
 & Reduction & Reduction & Increase & \\
\midrule
GPT-2-XL (1.5B) & 80.81\% & 80.17\% & 5.73\% & 6.798 \\
GPT-Neo (1.3B) & 80.42\% & 80.21\% & 7.08\% & 5.156 \\
\bottomrule
\end{tabular}
\end{table}

\subsubsection{Extraction Rates}

RMFT significantly reduces memorization relative to baseline fine-tuning on GPT-Neo. RMFT achieves an 80.42\% reduction in Total Extraction Rate (TER) and an 80.21\% reduction in Seen Extraction Rate (SER) compared to the baseline. These results are highly consistent with our GPT-2-XL findings (80.81\% TER reduction and 80.17\% SER reduction), demonstrating that RMFT's effectiveness generalizes across model architectures.

At the final checkpoint, RMFT achieves an 82.8\% reduction in TER and 82.87\% reduction in SER. These final checkpoint results are comparable to GPT-2-XL's 87.62\% TER reduction and 86.0\% SER reduction, showing consistent privacy protection across both models.

\begin{figure}[h]
\centering
\begin{subfigure}{0.48\textwidth}
    \includegraphics[width=\textwidth]{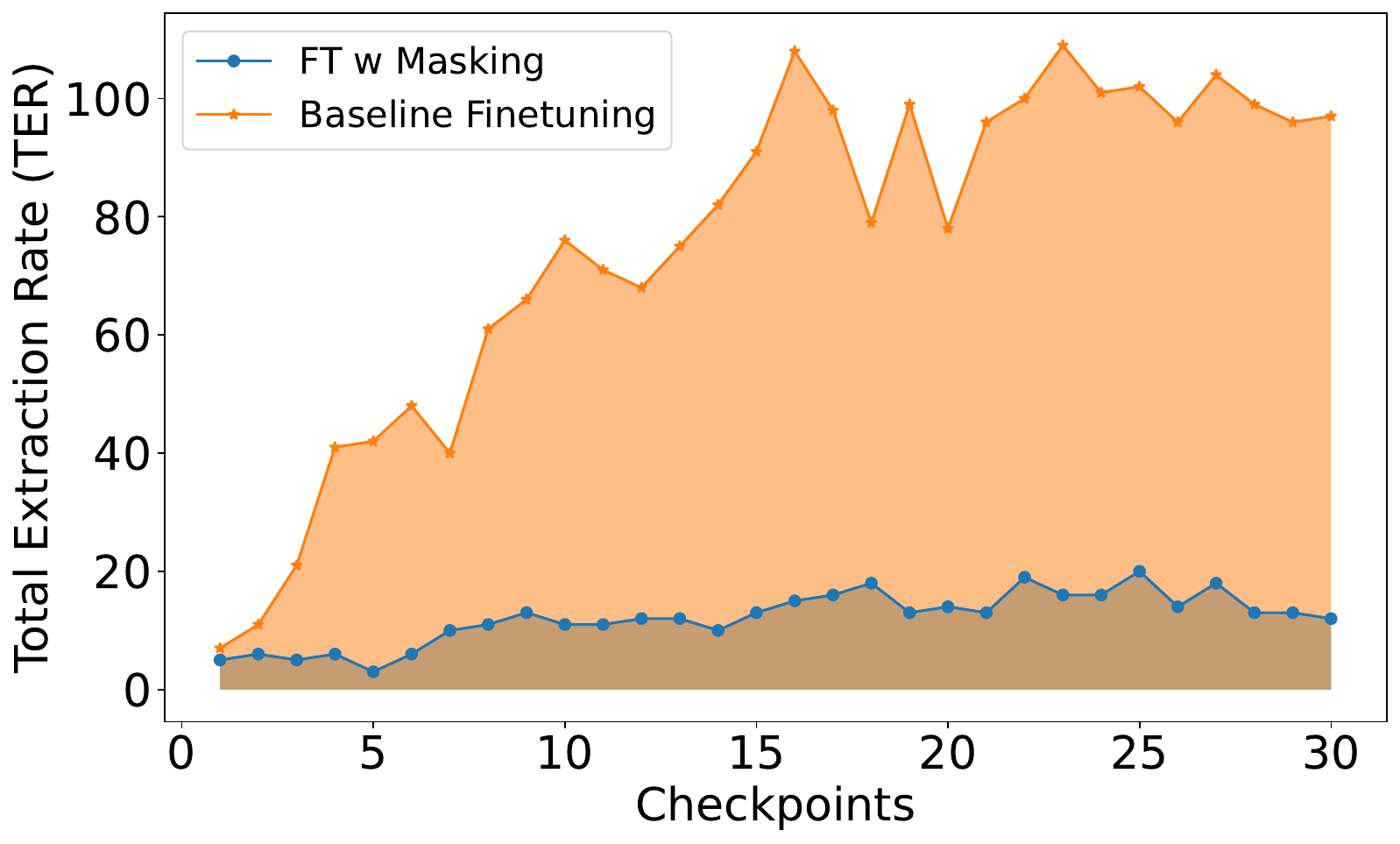}
    \caption{TER for GPT-2-XL}
    \label{fig:ter-gpt}
\end{subfigure}
\hfill
\begin{subfigure}{0.48\textwidth}
    \includegraphics[width=\textwidth]{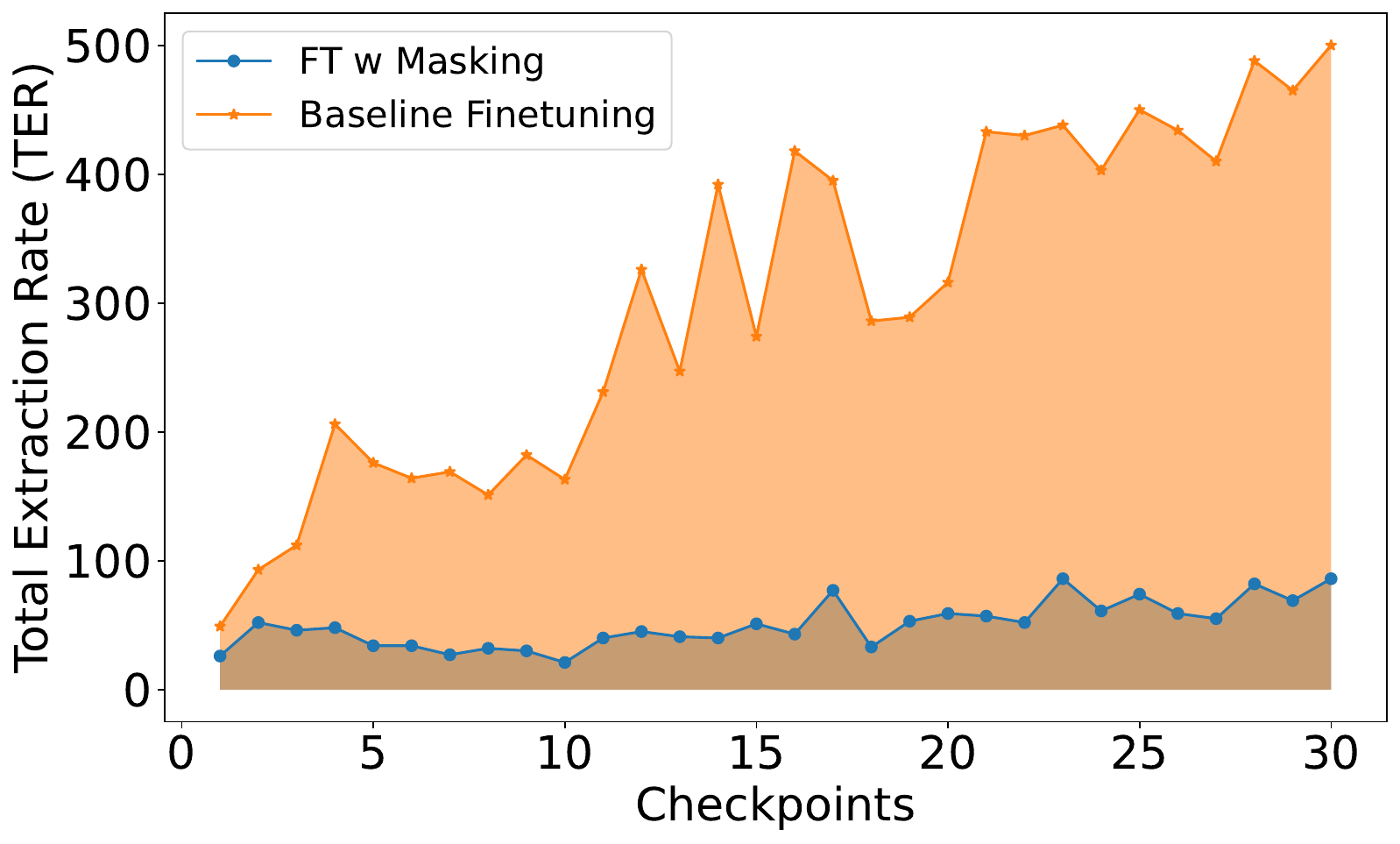}
    \caption{TER for GPT-Neo-1.3B}
    \label{fig:ter-neo}
\end{subfigure}

\vspace{0.3cm}

\begin{subfigure}{0.48\textwidth}
    \includegraphics[width=\textwidth]{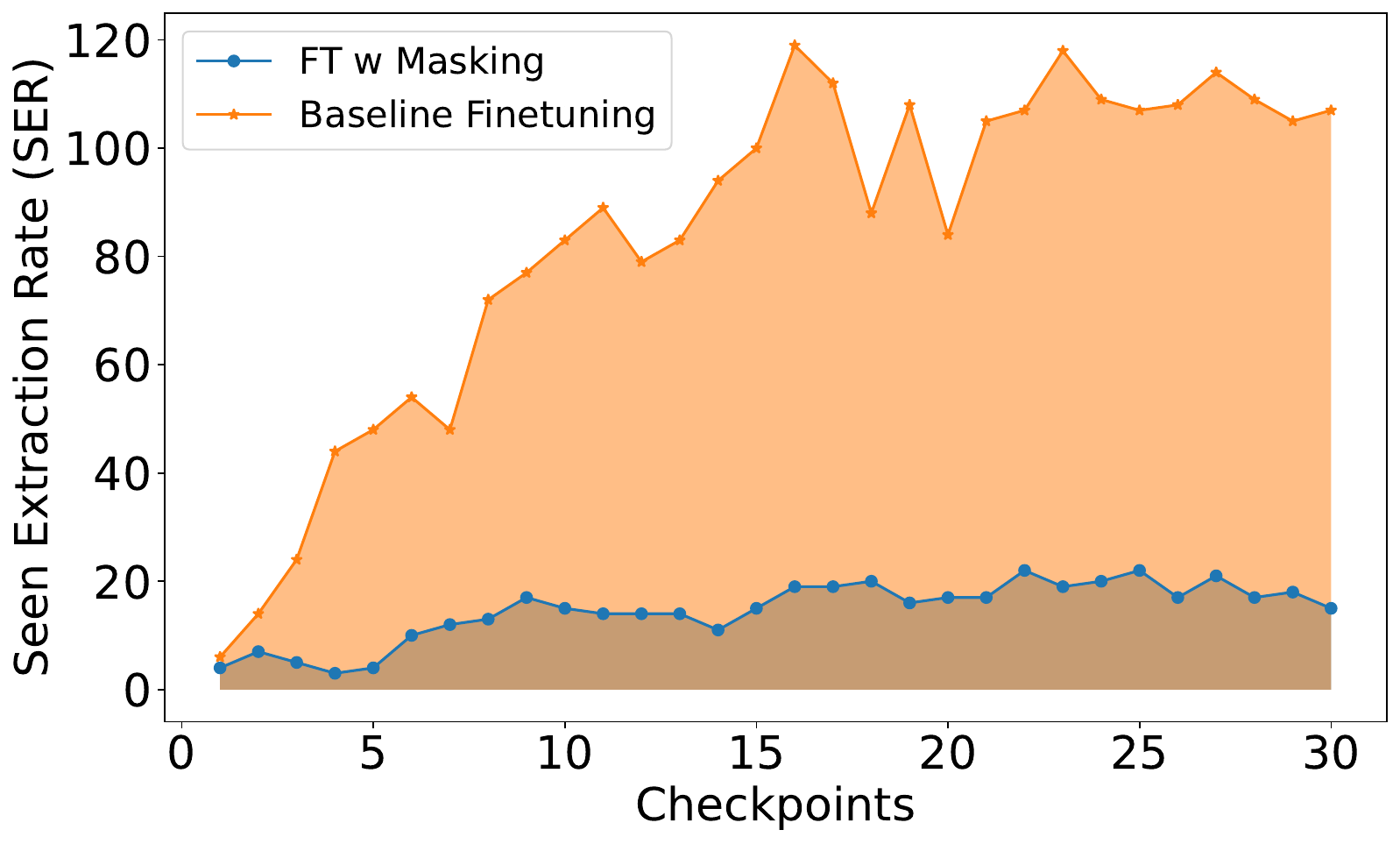}
    \caption{SER for GPT-2-XL}
    \label{fig:ser-gpt}
\end{subfigure}
\hfill
\begin{subfigure}{0.48\textwidth}
    \includegraphics[width=\textwidth]{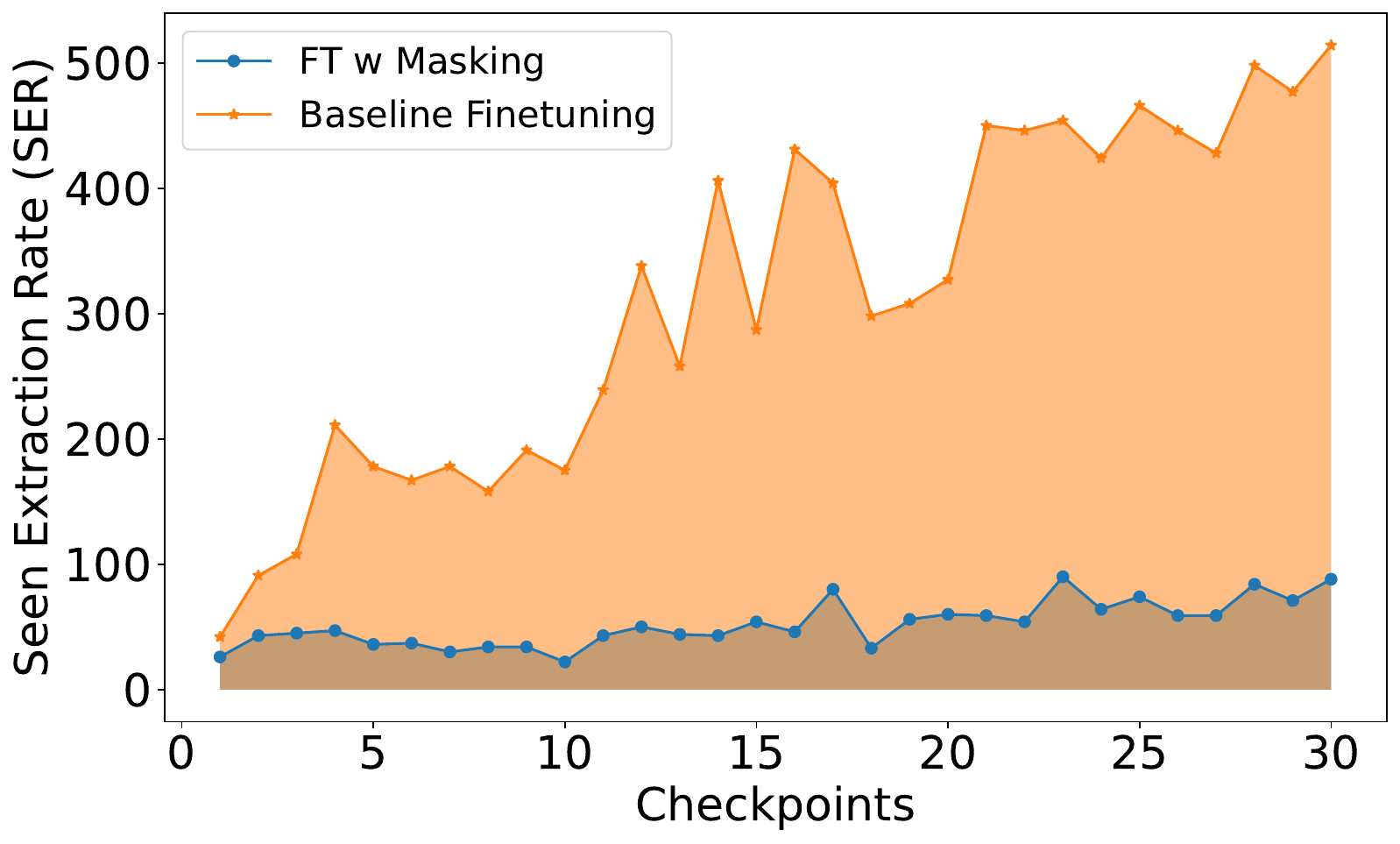}
    \caption{SER for GPT-Neo-1.3B}
    \label{fig:ser-neo}
\end{subfigure}

\caption{Comparison of extraction rates across training checkpoints for GPT-2-XL and GPT-Neo-1.3B. (a) Total Extraction Rate (TER) for GPT-2-XL, (b) TER for GPT-Neo-1.3B, (c) Seen Extraction Rate (SER) for GPT-2-XL, and (d) SER for GPT-Neo-1.3B. Both models exhibit similar trends, with RMFT consistently achieving lower extraction rates than baseline across all checkpoints. Lower values indicate fewer emails memorized.}
\label{fig:gptneo-comparison}
\end{figure}

Figure~\ref{fig:gptneo-comparison} compares the extraction rates across training checkpoints for both GPT-2-XL and GPT-Neo-1.3B.

\subsubsection{Mean Delta Perplexity}

RMFT achieves an average perplexity increase of 7.08\% relative to the baseline on GPT-Neo. While this is slightly higher than GPT-2-XL's 5.73\% increase, it remains within acceptable bounds for privacy-preserving fine-tuning. The consistency in memorization reduction (approx. 80\%) across both models, despite minor variations in perplexity impact, validates RMFT's architecture-agnostic design.

\begin{figure}[h]
\centering
\begin{subfigure}{0.48\textwidth}
    \includegraphics[width=\textwidth]{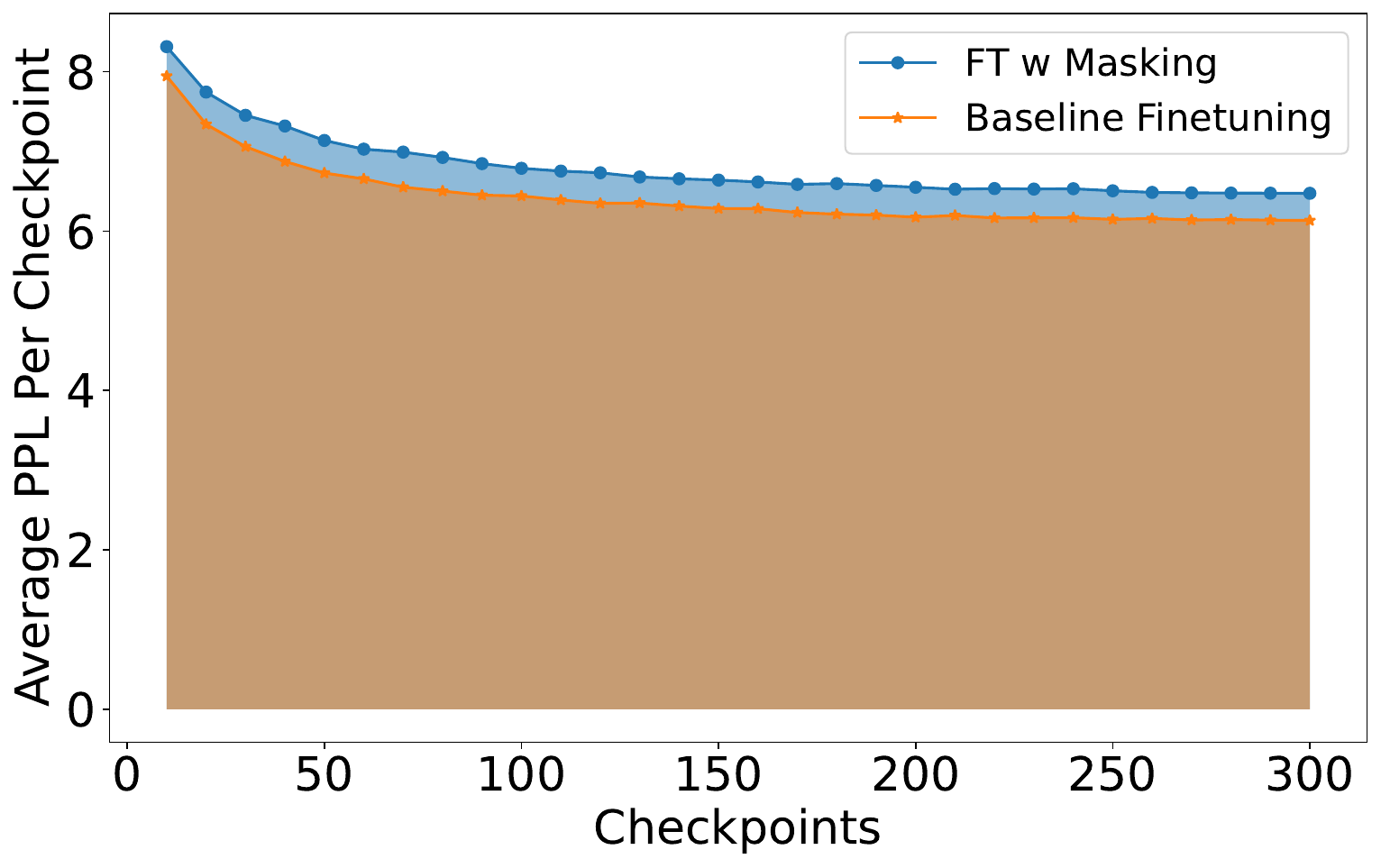}
    \caption{Average Perplexity for GPT-2-XL}
    \label{fig:avg-gpt-ppl}
\end{subfigure}
\hfill
\begin{subfigure}{0.48\textwidth}
    \includegraphics[width=\textwidth]{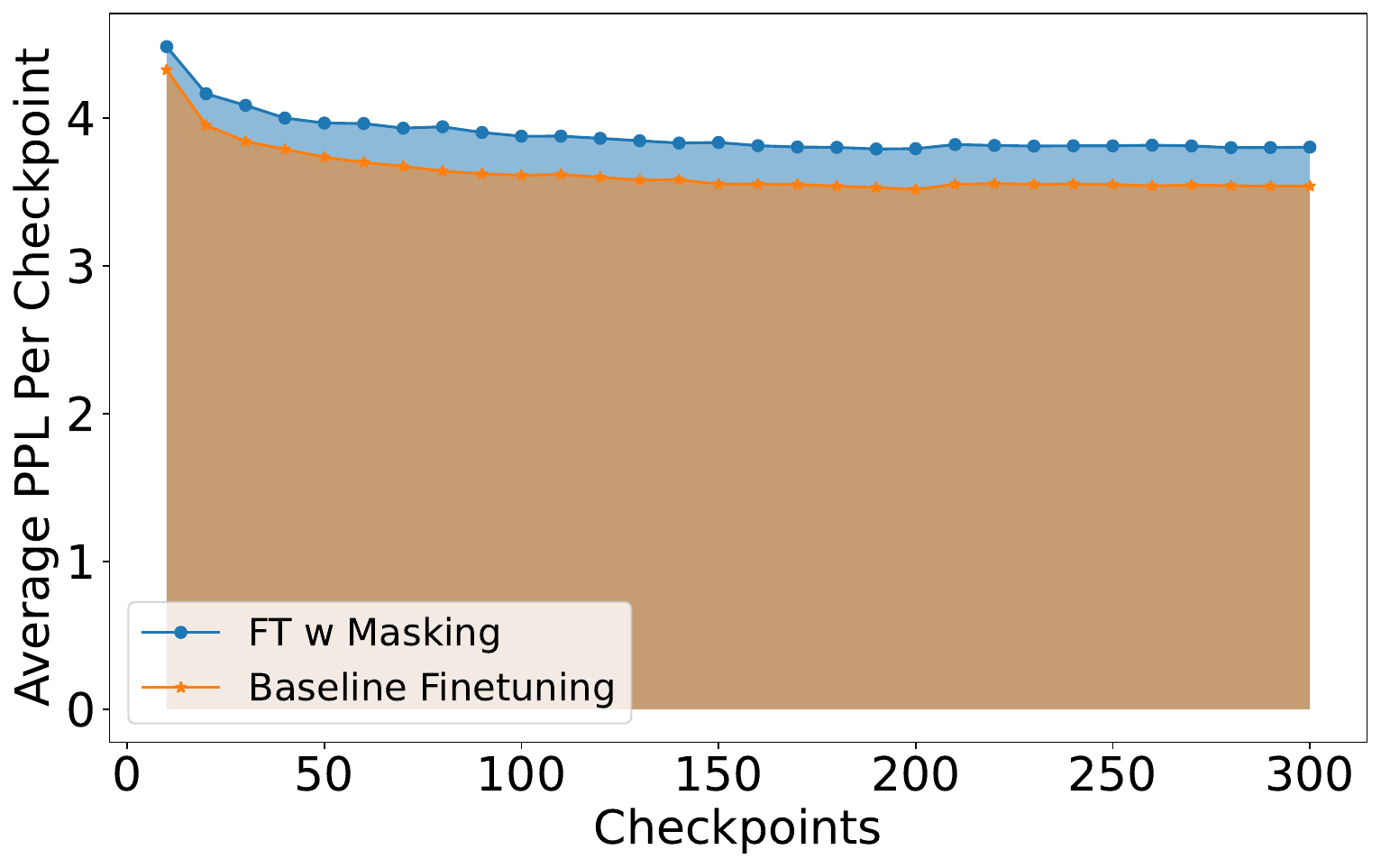}
    \caption{Average Perplexity for GPT-Neo-1.3B}
    \label{fig:avg-neo-ppl}
\end{subfigure}

\caption{Average perplexity per checkpoint for GPT-2-XL and GPT-Neo-1.3B across all training techniques. (a) GPT-2-XL shows RMFT maintains lower perplexity than deduplication throughout training. (b) GPT-Neo-1.3B exhibits similar trends, with RMFT achieving consistent performance across checkpoints. Both models demonstrate that RMFT preserves model quality better than deduplication.}
\label{fig:gptneo-perplexity}
\end{figure}

Figure~\ref{fig:gptneo-perplexity} compares the average perplexity per checkpoint for both models across all three training techniques.

\subsubsection{Summary}

These results demonstrate that RMFT's privacy-utility tradeoff generalizes across model scales and architectures. Both GPT-2-XL (1.5B) and GPT-Neo (1.3B) exhibit:
\begin{itemize}
    \item Consistent memorization reduction: approx. 80\% average TER and SER reduction
    \item Minimal performance impact:  less than 8\% perplexity increase
    \item Similar final checkpoint performance: greater than 82\% extraction reduction
\end{itemize}

This consistency supports RMFT's applicability to other transformer-based language models.

\end{document}